\pdfoutput=1

\documentclass[11pt]{article}

\usepackage[]{emnlp2021}

\usepackage{times}
\usepackage{latexsym}

\usepackage{graphicx}
\usepackage{import}
\usepackage{multirow}
\usepackage{booktabs}
\usepackage{longtable}
\usepackage{colortbl}

\usepackage{abbrevs}

\makeatletter
\renewcommand\maybe@space@{%
  \maybe@ictrue
  \expandafter   \@tfor
    \expandafter \reserved@a
    \expandafter :%
    \expandafter =%
                 \nospacelist
                 \do \t@st@ic
  \ifmaybe@ic
    \space
  \fi
}
\makeatother

\newabbrev{\UNUSED}{unused}[UNUSED]

\newabbrev{\ACC}{Automated Compliance Checking}[ACC]
\newabbrev{\BIM}{Building Information Model}[BIM]
\newabbrev{\NLP}{Natural Language Processing}[NLP]
\newabbrev{\RBS}{Rule-Based System}[RBS]
\newabbrev{\DL}{Deep Learning}[DL]
\newabbrev{\IFC}{Industry Foundation Classes}[IFC]
\newabbrev{\IR}{Information Retrieval}[IR]
\newabbrev{\IE}{Information Extraction}[IE]
\newabbrev{\MWE}{Multi-Word Expressions}[MWE]
\newabbrev{\NER}{Named Entity Recognition}[NER]
\newabbrev{\RC}{Relation Classification}[RC]
\newabbrev{\ML}{Machine Learning}[ML]
\newabbrev{\GRU}{Gated Recurrent Unit}[GRU]
\newabbrev{\POS}{Part-of-Speech}[POS]

\newabbrev{\NP}{Noun-Phrase}[NP]
\newabbrev{\VP}{Verb-Phrase}[VP]

\newabbrev{\biLSTM}{bidirectional Long Short-Term Memory}[bi-LSTM]
\newabbrev{\CRF}{Conditional Random Field}[CRF]
\newabbrev{\LM}{Language Modelling}[LM]

\newabbrev{\BERT}{\textsc{BERT}}

\usepackage[T1]{fontenc}

\usepackage[utf8]{inputenc}

\usepackage{microtype}

\title{SP{\large{A}}R.txt, a cheap Shallow Parsing approach for Regulatory texts}

\author{Ruben Kruiper$^{*}$ \and Ioannis Konstas  \and Alasdair Gray \\
    School of Mathematics and Computer Sciences \\
    Heriot-Watt University,
    Edinburgh, United Kingdom \\
  $^{*}$\texttt{ruben.kruiper@hw.ac.uk} \\\AND
  Farhad Sadeghineko \and Richard Watson \and Bimal Kumar \\
  Department of Architecture and Built Environment \\ 
  Northumbria University, 
  Newcastle, United Kingdom\\}

\begin{document}
\maketitle
\begin{abstract}
Automated Compliance Checking (ACC) systems aim to semantically parse building regulations to a set of rules. However, semantic parsing is known to be hard and requires large amounts of training data.
The complexity of creating such training data has led to research that focuses on small sub-tasks, such as shallow parsing or the extraction of a limited subset of rules.
This study introduces a shallow parsing task for which training data is relatively cheap to create, with the aim of learning a lexicon for ACC.
We annotate a small domain-specific dataset of 200 sentences, \textsc{SPaR}.txt\footnote{For the \textsc{ScotReg} corpus, \textsc{SPaR}.txt dataset and code see: \url{https://github.com/rubenkruiper/SPaR.txt}}, and train a sequence tagger that achieves 79,93 F1-score on the test set.
We then show through manual evaluation that the model identifies most (89,84\%) defined terms in a set of building regulation documents, and that both contiguous and discontiguous Multi-Word Expressions (MWE) are discovered with reasonable accuracy (70,3\%).
\end{abstract}

\section{Introduction}
Non-compliance with building regulations has been linked to fatal incidents \cite{Grenfell2017}. However, ensuring that a building complies with regulations is complicated and time-consuming because:
\begin{itemize}
    \item Regulations contain ambiguous and sometimes conflicting criteria
    \cite{Grenfell2017, Conflicts2020}.
    \item Regulations change and are distributed over many documents \cite{Fuchs2021}, e.g., over 800 documents in the U.K. with many of them behind a paywall.
    \item Criteria often refer to entire sections in other documents, e.g., ``\textit{The emergency lighting should be installed in accordance with BS 5266: Part 1: 2016 as read in association with BS 5266: Part 7: 1999 (BS EN: 1838: 2013).}'' \cite{ScotReg2020}.
    \item Regulations differ per country, and some criteria borrowed from international regulations are not suited to the specific environment \cite{Moon2019}.
\end{itemize}

\noindent
\ACC (\ACC) could reduce the difficulty, time, costs and number of human errors made during compliance checking \cite{Dimyadi2013, Preidel2018}, as well as ease customisation and innovation in the building sector \cite{Niemeijer2014}. There exist two branches of \ACC research. One focuses on tools that reason over a rule-base -- often consisting of hard-coded rules \cite{Pauwels2016, Solihin2019}. The other branch attempts semantically parse the Natural Language regulations into rules that enable reasoning -- a complex task that is of interest to the wider legal domain \cite{Wyner2012}.

This study presents a novel \textit{shallow parsing} task, for which the creation of training data is cheap, and an accompanying small dataset of 200 sentences. The aim is to learn a semantic lexicon for \ACC, which is often an important first step for semantic parsing because it enables the grounding of information units identified in a text -- such as objects, interactions and constraints \cite{Zettlemoyer2005, Kollar2010, Chen2012}. Section \ref{rw} motivates our task and describes related work with a focus on parsing building regulations. \ref{MWE_tagging} describes the task, as well as the collection of a small annotated dataset -- \textsc{SPaR}.txt -- for the task of discovering and identifying domain-specific terms, including {\MWE}s (\MWE). In \ref{comparison} we describe and train a sequence tagging model, which generalises well to unseen text within the same domain. \ref{evaluation} describes the evaluation of outputs, specifically with regards to the objects identified in a corpus of 13K sentences derived from the Scottish Building Regulations \cite{ScotReg2020}. Our model finds 89,84\% of the terms explicitly defined in these documents. Furthermore, a significant proportion (70,3\%) of these predictions matches exactly what a human annotator would consider to be an object in a given sentence context. We argue that our new task can provide a cheap approach to lexicon learning that could benefit (1) \IE (\IE) in support of \ACC, (2) \IR (\IR) in support of manual compliance checking, and (3) the mapping of unstructured text to a structured representation. 

\section{Related work} \label{rw}

\subsection{Semantic parsing for \ACC} \label{rw:semantic_parsing} 
Semantic parsing revolves around learning the meaning of Natural Language and converting it to an executable logical form, which is a hard and unsolved task \cite{Mooney2007, Artzi2013}. The fragmented structure of legal texts further complicates semantic parsing in legal domains \cite{Lawsky2017}, e.g., a clause may state that some object must comply with all of the regulations in some other section or document. Therefore, statutory reasoning requires defeasible logic representations that allow conclusions to be defeated on the basis of subsequent information \cite{Pertierra2017}. Due to the complexity of the task, existing approaches to semantic parsing in the \ACC domain often limit their scope to parsing quantitative requirements in some small sub-domain \cite{Fuchs2021, Moon2021}.

In the \ACC domain, studies on semantic parsing rely on traditional {\RBS}s to extract concepts and construct logical statements, e.g., \cite{ElGohary2016, Zhang2017, Zhou2019, Xu2019}. Empirical approaches may be able to handle the combinatorial explosion caused by ambiguity in Natural Language \cite{Wyner2012}. But training data for such systems is difficult and costly to collect \cite{Chen2012, Herzig2017}. Unsurprisingly, \ML studies in the \ACC domain focus on simpler sub-tasks of semantic parsing for which training data is relatively easy to collect, such as \NER \cite{Liu2017, Moon2021} and \RC \cite{Zhong2020}. This study follows a similar strategy to limit the complexity of annotation, see \ref{mwe:problem_statement} and \ref{mwe:task}.

\subsection{Shallow parsing for \ACC} \label{rw:shallow_parsing} 
Orthogonal research on shallow parsing in the \ACC domain includes (1) the decomposition of complex sentences into parts that are easier to process \cite{Zhang2019}, and (2) more-or-less idiosyncratic semantic markup schemes that help identify requirements and their components in text, e.g., \cite{Nisbet2011} and \cite{ElGohary2016}. Efforts to automate such shallow parsing approaches encountered performance issues when handling {\MWE}s \cite{Zhang2019, Zhang2020}. 

Although the proper handling of {\MWE}s is a key issue in \NLP \cite{Siskind1996, Sag2002, Ramisch2018}, extracting {\MWE}s is especially relevant to \IE in domains rich in technical terms \cite{Baldwin2010} -- such as the building regulations. Processing {\MWE}s is a general requirement for \ACC, because both single and multi-word concepts mentioned in regulations have to be aligned with components and values found in \BIM  (\BIM) models. Such \BIM models rely on standards, such as the \IFC (\IFC) data model, to facilitate amongst others compliance checking and information exchange between applications and potentially international stake-holders \cite{Plume2007, Beetz2009, Pauwels2016}. This study aims to automatically learn a vocabulary for \ACC, which entails {\MWE} processing.

\subsection{Processing {\MWE}s} \label{rw:mwe_identification} 
Processing \MWE can be divided into two  sub-tasks: \MWE discovery and \MWE identification \cite{Constant2017}. 

\MWE discovery aims to find new types of {\MWE}s in text corpora and storing them in a lexicon. Unsupervised approaches are used that rely on properties of {\MWE}s that set them apart from random combinations of words, such as word collocation frequency \cite{Manning2002, Pecina2006}, non-substitutability of component words \cite{Lapata2003, Constant2017}, and non-compositionality \cite{Frege1884, Riedl2015}. The latter applies mostly to idiomatic expressions\cite{Villavicencio2019}, such as `\textit{cloud nine}'. 

\begin{table*}[h!t]
    \centering \small
    \begin{tabular}{lrrr}
                                                                            & \textbf{Domestic} & \textbf{Non-Domestic} & \textbf{Total}   \\ \toprule
    Terms defined in definitions section                                    & 128       & 127         & 128     \\
    Defined terms in text after lemmatisation and lower-casing              & 233       & 247         & 292     \\
    Number of terms linking to definitions section                          & 4.687     & 5.368       & 10.055  \\ \arrayrulecolor{lightgray}\hline\arrayrulecolor{black}
    Number of tokens                                                        & 131.666   & 151.499     & 283.165 \\
    Vocabulary                                                              & 8.285     & 8.925       & 9.837   \\
    Number of sentences                                                     & 6.313     & 7.293       & 13.606  \\
    Mean sentence (word-level token) length, excluding punctuation                                            & 20,86     & 20,77       & 20,81  \\
    Standard deviation                                                      & 11,96     & 12,32       & 12,16 \\ \bottomrule
    \end{tabular}
    \caption{Statistics for the \textsc{ScotReg} corpus -- the number of defined terms, word-level tokens and sentences found in the domestic and non-domestic Scottish Building regulations.
    }
    \label{tab:regulations_tokens_sents}
\end{table*}

\MWE identification revolves around annotating {\MWE}s in a corpus, based on a lexicon or on results from \MWE discovery \cite{Constant2017}. This enables representing {\MWE}s as single tokens, which has been shown to improve accuracy of NLP tasks \cite{Green2011}, such as dependency parsing \cite{Nivre2004}. 
Supervised approaches are used, amongst which sequence tagging has been found to work well \cite{Constant2017}, e.g., \cite{Blunsom2006, Constant2012}. Sequence tagging has also been used for joint \MWE identification and \POS (\POS) tagging \cite{Constant2011} and may be amended to handle discontiguous {\MWE}s \cite{Schneider2014}. 

This study explores sequence tagging for joint \MWE processing. However, this study does not aim to handle idiomatic expressions or proverbs. Beyond research on \MWE processing, a related task that focuses on identifying technical terms and Named Entities is concept mining, e.g., \cite{Rajagopal2013, Poria}. In contrast to these concept mining studies, we do not rely on dependency parses or external resources, such as ConceptNet \cite{Speer2013}. 

\section{\MWE tagging for \ACC} \label{MWE_tagging} 

\subsection{A building regulations corpus} \label{mwe:data_collection} 
Many building regulations are captured in formats that are not easily processed by computers, such as PDF \cite{Fuchs2021}. The Scottish Building Regulations \cite{ScotReg2020} are an exception and are openly available online. We scrape the domestic and non-domestic regulations, including text found in lists, side-notes, tables and captions. 

We use TextBlob \footnote{\url{https://textblob.readthedocs.io/en/dev/}} for word-level tokenization (Penn Treebank Tokenizer) and sentence splitting (PunktSentenceTokenizer \cite{Strunk2006}).
We will refer to the resulting corpus as \textsc{ScotReg}. An overview of the size of \textsc{ScotReg} in terms of sentences and tokens can be found in Table~\ref{tab:regulations_tokens_sents},
as well as the number the terms defined in the `\textit{Terms and definitions}' sections. Throughout \textsc{ScotReg} defined terms provide a hyperlink to the definitions section. This enables us to count how often defined terms occur. Note these defined terms represent \textit{classes}, they are expressed in various ways throughout the texts and some of the variations do not match any of the terms verbatim -- even after lower-casing and lemmatizing.


\subsection{Problem statement} \label{mwe:problem_statement} 
The \IFC schema does not comprehensively cover the terminology used in the building regulations. In ``\textit{A roof covering or roof light which forms part of an internal ceiling lining should} [...]''\footnote{For an overview of used examples, the full sentences, and origin of these examples see Table~\ref{tab:examples_overview} in Appendix~\ref{examples_overviewe}.}
the term `\textit{roof light}' would fall under the more generic \IFC class `\textit{window}'. The problem is compounded as building regulations cover a wide range topics (and thus concepts) -- from design and construction, including fire regulations and accessibility, to facility management, renovation and demolition \cite{Pauwels2016}. Despite the more fine-grained terminology used in the \textsc{ScotReg} corpus, only 128 terms are defined -- neither `\textit{roof light}', `\textit{roof covering}' nor `\textit{internal ceiling lining}' are defined.

\subsection{Task description} \label{mwe:task} 
The task in this study is to identify low-level constituent parts of a sentence, which we will refer to as \textit{spans}. We assume that contextual inter-dependencies between single-word and multi-word spans help tackle lexical sparsity of {\MWE}s. Therefore, we tag both single words and groups of words in a sentence and we tag all tokens exhaustively, including punctuation. Whether spans comprise a single punctuation mark or a group of possibly discontiguous tokens, they should represent a coherent unit of grammatical meaning. 
\begin{itemize}
    \item ``\textit{The Building (Scotland) Act 2003 gives Scottish Ministers} [...]''
\end{itemize}
\noindent
`\textit{The Building (Scotland) Act 2003}' is a Named Entity 
that should be treated as a single span. But a regular noun-chunking approach would break on the parentheses and the cardinal number -- it may even treat every capitalised word as a proper noun.
\begin{itemize}
    \item ``\textit{A roof covering or roof light which forms part of an internal ceiling lining should} [...]''
\end{itemize} 
\noindent
A constituency parser may break the term `\textit{roof covering}' into separate words, because a \POS tagger would typically assign  `\textit{covering}' the tag \textsc{VBG}.

\begin{itemize}
    \item ``[...]\textit{, a paved (or equivalent) footpath at least 900mm wide} [...]''
\end{itemize}
\noindent
In this case, while we would like to split `\textit{a paved (or equivalent) footpath}' into the spans: `\textit{a paved footpath}', `\textit{(}', `\textit{)}', `\textit{or}' and `\textit{equivalent}'. In a downstream task this would allow us to define a class for `\textit{paved footpath}', and reason over the equivalent types of footpaths. However, to extract `\textit{a paved footpath}' from the sentence above, a discontiguous span representation is required.


\begin{figure}[t]
    \centering 
    \includegraphics[width=\linewidth]{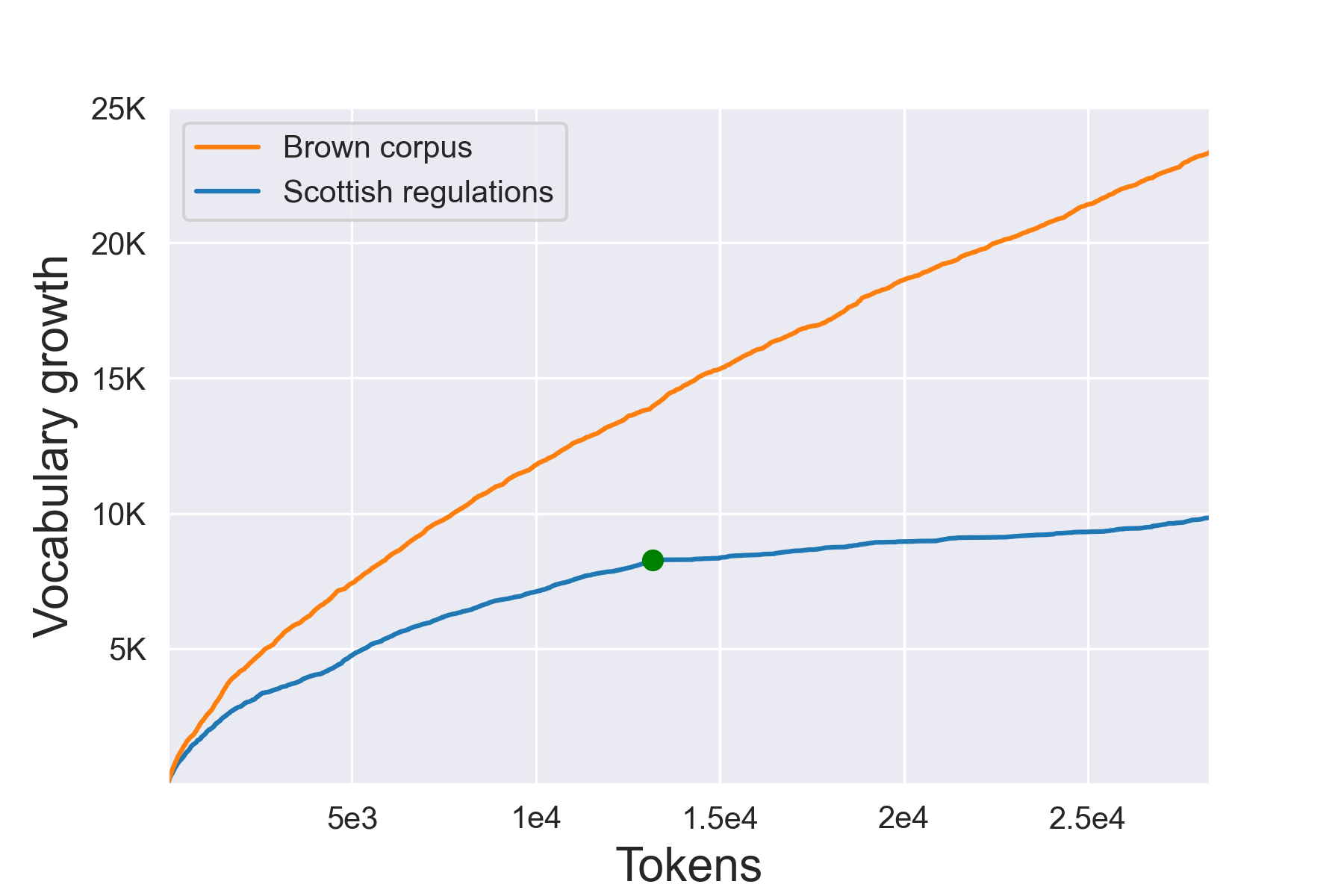}
    \caption{Vocabulary growth for the \textsc{ScotReg} corpus in comparison to the heterogeneous \textsc{Brown} corpus. The dot indicates the division between domestic and non-domestic building regulations.}
    \label{fig:vocab_growth}
\end{figure}

\subsection{Simplifying assumptions} \label{mwe:assumptions}
For our task we rely on a simplified definition of {\MWE}s: ``\textit{possibly discontiguous combinations of at least two tokens, where tokens are separated by white-spaces or punctuation in text}'' -- similar to \cite{Villavicencio2019}. We also make two assumptions on the types of {\MWE}s that we expect to find in the building regulation texts. First, we assume that the building regulations contain few to no idiomatic expressions, because these may introduce ambiguity. We justify this assumption as the building regulations use a slightly more formal syntax, albeit not as strict as other types of legal text \cite{Ilias2020}.
\begin{figure*}[ht]
    \centering 
    \includegraphics[width=.9\linewidth]{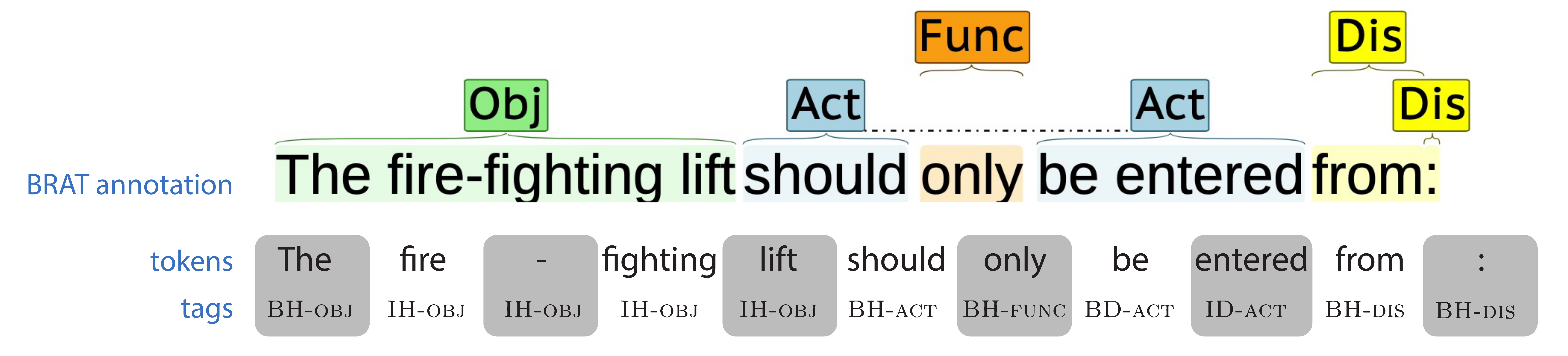}
    \caption{Example of an annotated sentence. The determiner at the start of the \textsc{Object} span is taken to be part of the span. A discontiguous \textsc{Action} span is interjected by a \textsc{Functional} span that modifies the Verb-Phrase. During training the sentence is tokenized and the aim is to predict the correct tags for each token, see the tagging scheme described in Section~\ref{training:tags}. The identifier for this sentence in the dataset is `\textit{d\_2.14.4\_i3\_s\_0}'.}
    \label{fig:annotation_example}
\end{figure*}

Second, we expect a relatively low variability in surface forms. Both for verbal expressions found in clauses, e.g., `\textit{X should conform to Y}, and for the surface forms of {\MWE}s that indicate technical terms, e.g., `\textit{insulation envelope}' and `\textit{self-contained emergency luminaries}'.
A low variability in surface forms would be reflected by a relatively small vocabulary -- which is thought to ease the complexity of various \NLP tasks \cite{Church2013HowKnow}. As can be seen in Figure~\ref{fig:vocab_growth}, \textsc{ScotReg} has in the order of 10K unique tokens for a total of 283K. In comparison to the more heterogeneous Brown corpus \cite{Brown1964}, which has more than 23K unique tokens for the first 283K tokens, \textsc{ScotReg} indeed has a small vocabulary.

\subsection{Annotating SP{\small{A}}R.txt} \label{mwe:data}

A domain expert annotated a random selection of 200 sentences in \textsc{BRAT} \cite{Stenetorp2012}. Our assumption is that such a small dataset should suffice for achieving reasonable results on the proposed parsing task. Figure~\ref{fig:annotation_example} exemplifies how annotations can span single words, multiple words and also indicate that two groups of words belong to a single, discontiguous span. To distinguish between verb-based and noun-based spans, as well as spans that belong to neither of these classes, we annotate the following span types:
\begin{itemize}
    \item \textsc{Object} spans indicate either real-world objects or distinguishable concepts. They include proper nouns, compounds, multi-word terms, and multi-word Named Entities, such as `\textit{the Target Emissions Rating}', `\textit{offensive fire-fighting}' and `\textit{BS 8000-15: 1990}'. We include determiners as part of the \textsc{Object} span during annotation, see Figure~\ref{fig:annotation_example}.
    \item  \textsc{Action} spans may help identify whether a sentence expresses a requirement, similar to  \cite{Nisbet2011}. We include verbs, support verbs, prepositional verbs and verb-particle constructions, e.g., `\textit{should be maintained}' and `\textit{takes account of}', but we expect to split light-verb constructions, such as `\textit{to take a shower}' \cite{Constant2017} into `\textit{to take}' and the \textsc{Object} span `\textit{a shower}'. 
    \item  \textsc{Functional} spans are modifiers that are not inherently part of an \textsc{Object} or \textsc{Action} span. They include adverbs and adjectives, e.g.,  `\textit{main}' and `\textit{principal}' in `\textit{the main or principal bedroom}', as well as complex function words, e.g., `\textit{up to}'.
    \item \textsc{Discourse} spans include punctuation, co-reference anaphora, conjunctions and disjunctions, e.g., `\textit{,}', `\textit{or}' and `\textit{this}'.
\end{itemize}

\noindent 
{\MWE}s can blur the lines between syntax and semantics \cite{Green2011}, with ambiguous cases leading to annotation inconsistencies \cite{Hollenstein2016}. Because the task involves annotating both the words that make up an \MWE and those surrounding it, the annotator is forced to come up with more-or-less consistent decisions. As a tool to determine which words belong together, annotators are asked to rely on standard constituency tests:
\begin{itemize}
    \item Substitution test --  if you can substitute a part of a sentence with another word or group of words that belong to the same type, the part is a constituent.
    \item Pronoun test -- if you can replace a part of a sentence with `it’, the part is a constituent.
    \item Question by repetition test -- if you can repeat a part of a sentence, within a valid question, then it is a constituent.
\end{itemize}

\begin{figure}[ht]
    \centering 
    \includegraphics[width=1\linewidth]{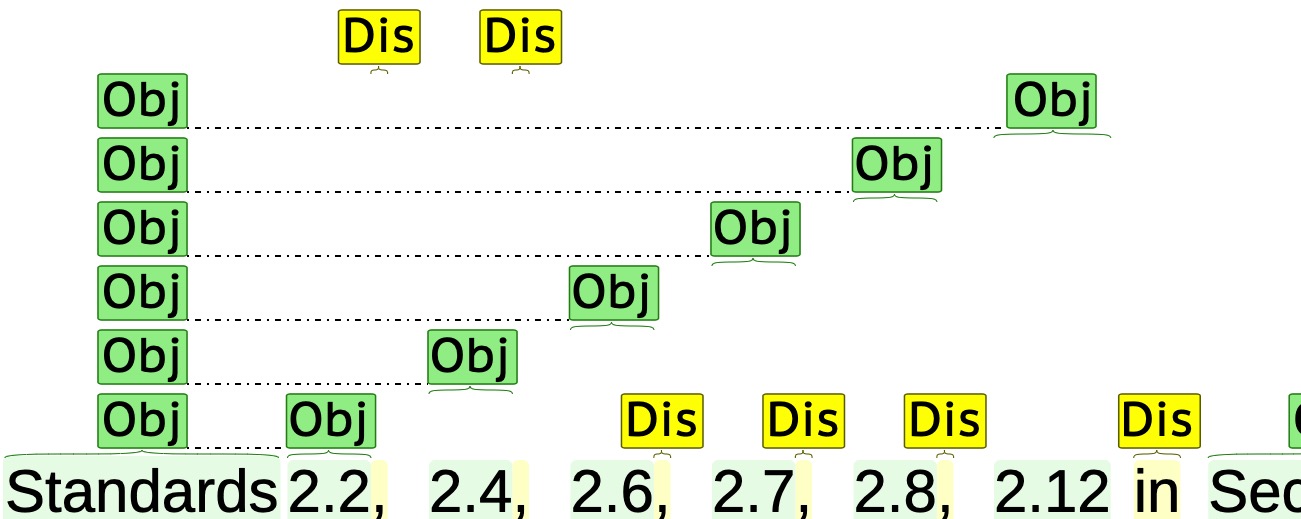}
    \caption{Example of an annotated coordination, where each of the items shares the word `\textit{Standards}'. Note that only part of the sentence is shown here, the identifier for this sentence in the dataset is `\textit{d\_0.12.2\_i3\_\#1\_s\_0}'.}
    \label{fig:list_example}
\end{figure}

\noindent
Nevertheless, annotating {\MWE}s often requires domain-specific knowledge and remains ambiguous. Using Figure~\ref{fig:annotation_example} as an example, one might argue that the preposition `\textit{from}' could be part of the \textsc{Action} span `\textit{should be entered}' -- considering that `\textit{from}' converts `\textit{to enter}' from an intransitive to a transitive verb. We define several loose guidelines to warrant further consistency in annotation:
\begin{enumerate}
    \item \textbf{Punctuation} Unless punctuation should be part of an \textsc{Object} span, such as the colon in the document name `\textit{BS 800-15:1990}', then punctuation should be marked as \textsc{discourse}. 
    \item \textbf{Negation} Wherever negation is separable from other spans, e.g., `[...] \textit{is not level} [...]', it should be annotated separately as a \textsc{Functional} span. In cases where negation is not separable, e.g., in `[...] \textit{cannot gain} [...]', the current approach is to tag the word containing the negation as \textsc{Functional}.
    \item \textbf{Granularity} Annotators should rely on domain knowledge to determine whether a \NP should be split, e.g., `[...] \textit{the effect of smoke travelling along a ceiling} [...]' should be broken up while ``[...] \textit{fire and rescue service personnel} [...]' should form a single span.
    \item \textbf{Coordination} Coordinating conjunctions and disjunctions should be split. Coordinated {\MWE}s may share a span, e.g., Figure~\ref{fig:list_example} shows how a list of `\textit{Standards}' share a word that is crucial to the semantics of the individual items. Similarly, conjunctions and disjunctions sometimes share a determiner, e.g., `\textit{the size and orientation of the windows}' would include the spans `\textit{the size}' and `\textit{the orientation}'. The motivation is to help downstream tasks determine that both `\textit{size}' and `\textit{orientation}', here, are properties of `\textit{the window}'. 
    \item \textbf{Overlap}  We currently limit the annotation of discontiguous spans to a maximum of two parts. Therefore, overlapping spans should only occur this cannot be avoided, e.g., `\textit{the Silver level}' and `\textit{the Gold level}' in `\textit{the Silver and Gold level}' would need to exist of three single-word spans to avoid overlap.
    \item \textbf{Adverbs} typically express a manner, place, time, frequency, degree, and so on. The expectation is that such expressions will usually be labelled as \textsc{Functional}, e.g., `\textit{only}' in Figure~\ref{fig:annotation_example}.
    \item \textbf{Adjectives} typically modify nouns and may or may not be part of an \textsc{Object} span. This decision would be based on whether the modified noun is likely to constitute a separate category or not. As an example, `\textit{structural}' would be part of the \textsc{Object} span `\textit{the structural properties}' but a separate \textsc{Functional} span in `\textit{matters of structural concern}'.
    \item \textbf{Quantities and units} are treated as a single \textsc{Object} span. If such a span modifies a noun, e.g., `\textit{900mm wide}', then these would usually form two separate spans `\textit{900mm}' and `\textit{wide}'.
\end{enumerate}

\noindent 
A second domain expert annotated 140 out of 200 sentences. The inter-annotator agreement was found to be Cohen \textit{k}=0,79. We randomly divide the gold annotations into a 60/20/20 (\%) split for train, development and test respectively 
-- see Table~\ref{tab:overview_spans} in Appendix \ref{appendix}.

\section{Training} \label{comparison}
\subsection{Representing discontiguous spans} \label{training:tags}
Regular tagging schemes, such as \textsc{BIO} and \textsc{BIOUL}, are unable to represent discontiguous spans \cite{Schneider2014}. We adopt a tagging approach that can handle discontiguous pairs of spans, similar to \cite{Muis2016}. Specifically, we use \textsc{BH} to indicate the beginning (head) of a span and \textsc{IH} to indicate subsequent tokens that belong to this span. If a second span exists that is part of a discontiguous \MWE, then the beginning (discontiguous) of this second span is tagged \textsc{BD}, and subsequent tokens of that span are tagged \textsc{ID}. Tags are provided with a type to distinguish between \textsc{Object} and \textsc{Action} spans etc. Figure~\ref{fig:annotation_example} provides example tags for an annotated sentence. Notably, we assume that between the head and discontiguous spans of a given type, there exist no head-spans of the same type.

\subsection{Model}\label{c:model} 
We adopt a sequence tagging approach that has been shown to work well \cite{Huang2015}, where an embedded text sequence is encoded by a \biLSTM network \cite{Hochreiter1997} and a \CRF \cite{Lafferty1999} model is used to predict a tag for each token in the sequence. We modify the implementation found in AllenNLP\footnote{\url{https://github.com/allenai/allennlp-models}} \cite{AllenNLP}. We embed text using pre-trained \textsc{BERT} embeddings (\textit{bert-base-cased}) \cite{Devlin2018} and rely on a single \biLSTM layer (hidden dim. 384). We add a self-attention layer that captures relative positional information \cite{Shaw2018}, following \cite{Huang2019}. The encoded and attended representations are concatenated and projected through two linear layers (hidden dim. 60) before being passed to the \CRF model. Table~\ref{tab:ablations} provides an overview of results.

We compare the use of \textsc{BERT} \cite{Devlin2018} and \textsc{SpanBERT} (\textit{SpanBERT/spanbert-base-cased}) \cite{Joshi2020} for tokenization and embedding. Our intuition is that the masked span \LM task may help capture {\MWE} properties used for \MWE discovery, see \ref{rw:mwe_identification}. But in our experiments \textsc{SpanBERT} embeddings consistently perform worse, see Table~\ref{tab:ablations}. This may be due to a mismatch between the span types and sizes that \textsc{SpanBERT} was originally trained on --  in \textsc{SpanBERT} all masked spans are contiguous -- and the ones found in our training dataset.
A more general issue is the imbalance of tag-types, see Table~\ref{tab:test_results_per_tag} for results on the test set per tag type.

\section{Evaluation} \label{evaluation}
Evaluating discovered lexical entries can be tricky \cite{Constant2017}. We limit our evaluation to the processing of \textsc{Object} spans. First, we perform an extrinsic evaluation by comparing our model output against the defined terms in the \textsc{ScotReg} corpus. Second, post hoc human judgement provides insight in the number of correctly identified \textsc{Object} spans given the sentence context.

\begin{table}[t] 
    \centering \small
    \begin{tabular}{llccc}
             & & \textbf{P}     & \textbf{R}     & \textbf{F1}    \\ \toprule
    development & BERT      & \textbf{80,18} & \textbf{81,76} & \textbf{80,96} \\
                & SpanBERT  & 75,54 & 79,33 & 77,39 \\ \arrayrulecolor{lightgray}\hline\arrayrulecolor{black}
    test        & BERT      & \textbf{79,93} & \textbf{80,89} & \textbf{79,93} \\
                & SpanBERT  & 75,37 & 78,10 & 76,48 \\ \bottomrule          
    \end{tabular}
    \caption{Precision (P), recall (R), and F1-score (F1) on the development and test sets.}
    \label{tab:ablations}
\end{table}{}

\begin{table}[t]
\centering \small
\begin{tabular}{lrrrr}
        & \multicolumn{1}{c}{\textbf{P}}    & \multicolumn{1}{c}{\textbf{R}}    & \multicolumn{1}{c}{\textbf{F1}}   & \multicolumn{1}{c}{\textbf{support}} \\ \toprule
BH-obj       & 80,68     & 86,53  & 83,50    & 193     \\
IH-obj       & 86,73     & 89,79  & 88,24    & 284     \\
BD-obj       & 60,00     & 40,00  & 48,00    & 30      \\
ID-obj       & 0,00      & 0,00   & 0,00     & 15      \\ \arrayrulecolor{lightgray}\hline
BH-act       & 76,15     & 83,84  & 79,81    & 99      \\
IH-act       & 71,21     & 69,12  & 70,15    & 68      \\
BD-act       & 50,00     & 37,50  & 42,86    & 8       \\
ID-act       & 0,00      & 0,00   & 0,00     & 4       \\ \hline
BH-dis       & 89,74     & 92,11  & 90,91    & 266     \\
IH-dis       & 100,00    & 31,82  & 48,28    & 22      \\\hline
BH-func      & 65,62     & 63,64  & 64,62    & 66      \\
IH-func      & 31,82     & 38,89  & 35,00    & 18      \\ \hline
             &    &   &     &         \\
accuracy     &           &         & 80,89    & 1.073    \\
macro avg.    & 59,33     & 52,77  & 54,28    & 1.073    \\
weighted avg. & 79,93     & 80,89  & 79,93    & 1.073   \\ \arrayrulecolor{black}\bottomrule
\end{tabular}
 \caption{Overview of results on the test set, broken down per tag type.}
    \label{tab:test_results_per_tag}
\end{table}

\subsection{Predicting defined terms} \label{e:defined_terms}
We use the model trained on \textsc{SPaR}.txt to predict tags for all sentences in the \textsc{ScotReg} corpus. We then compare whether each of the 128 defined terms is identified at least once by our model. A total of 115 (89,84\%) defined terms were found. Table~\ref{tab:defined_terms_found} lists the 13 defined terms (10,16\%) that were not found. Most of these defined terms exist verbatim in the \textsc{ScotReg} corpus, but our model splits these spans into multiple parts. However:
\begin{itemize}
    \item `\textit{average flush}' only occurs as `\textit{average flush volume}' and our model treats `\textit{average}' as a separate \textsc{Functional} span. 
    \item `\textit{High-speed ready in-building physical infrastructure}' never occurs verbatim in the text. 
\end{itemize}

\begin{table*}[t]
\centering \small
    \begin{tabular}{p{0.25\linewidth}p{0.65\linewidth}} 
    \textbf{Undetected defined terms} & \textbf{What we extracted} \\ \toprule
    Alternative exit                                         & Probably split, we find `\textit{alternative}' and `\textit{exit}'   \\ \arrayrulecolor{lightgray}\hline
    Average flush                                            & Probably split, we find `\textit{flush}' and several related terms, e.g., `\textit{dual flush type}' and `\textit{flush volume}'      \\ \hline
    Different occupation                                     & Probably split, we find `\textit{land}' and various types of `\textit{occupation}', e.g., `\textit{single occupation}', `\textit{multiple occupation}', `\textit{communal occupation}'                              \\ \hline
    Land in different occupation                             & See above \\ \hline
    Factory (class 1)                                      & Probably split, we find `\textit{factory}', `\textit{factory building}' and their plurals  \\ \hline
    Factory (class 2)                                      & See above, we do find `\textit{'factory class 2'}' \\ \hline
    High-speed ready in-building physical infrastructure & We find `\textit{high-speed-ready in-building physical infrastructure}'   \\ \hline
    Major renovation works                                   & Probably split, we find `\textit{renovation}' and various types of `\textit{works}', e.g., `\textit{chemical works}', `\textit{sewage work}' and `\textit{protective works}' \\ \hline
    Place of special fire risk                               & Probably split, we find the discontiguous span `\textit{place}' `\textit{fire risk}'  \\ \hline
    Public open space                                        & Not found, we find various types of `\textit{space}', e.g., `\textit{clear space}', `\textit{outdoor space}' and `\textit{communal spaces}'\\ \hline
    Reasonably practicable                                   & Probably split, we find `\textit{practicable}'  \\ \hline
    Storage building (class 1)                               & Probably split, we find `\textit{storage building}' and `\textit{storage buildings}' \\ \hline
    Storage building (class 2)                               & See above          \\ \arrayrulecolor{black}\bottomrule
    \end{tabular}
     \caption{Overview of the 13 defined terms that were not found in the text by our trained tagger.}
    \label{tab:defined_terms_found}
\end{table*}

\subsection{Post hoc human judgement} \label{e:objects}
We collect all contiguous and discontiguous \textsc{Object} spans that our model predicted on the 13K sentences of the \textsc{ScotReg} corpus. A total of 16,428 unique potential \textsc{Object} spans are identified. We find that this number decreases slightly, to 15,662, if we remove determiners and lower-case the text. We randomly select 165 out of the 16,5K \textsc{Object} spans, and then one of the sentences in which this object occurs -- this sample size provides a 99\% confidence level with a 10\% margin of error. We exclude objects that match any of the defined terms and exclude sentences that are part of the annotated dataset. 

We use Doccano \cite{Doccano2018} for annotation. Each of these 165 samples is presented to the annotators as a combination of the \textsc{Object} span and the corresponding sentence context. The task is to annotate whether the predicted \textsc{Object} span is actually an object in the sentence, with a choice between the labels: (1) exact match, (2) partial match, or (3) not an object. 

\begin{table}[ht]
\centering \small
    \begin{tabular}{llll}
                & \multicolumn{2}{c}{Object}   &               \\
                & Exact match  & Partial match & Not an object \\ \toprule
    Ann. 1 & 114 (69,1\%) & 41 (24,8\%)   & 10 (6,1\%)    \\
    Ann. 2 & 118 (71,5\%) & 39 (23,6\%)   & 8 (4,8\%)     \\ \arrayrulecolor{lightgray}\hline\arrayrulecolor{black}
    \textbf{Avg. }   & \textbf{116 (70,3\%)} & \textbf{40 (24,2\%) }  & \textbf{9 (5,5\%) }   \\ \bottomrule
    \end{tabular}
     \caption{Overview of labels by two annotators on 165 \textsc{Object} spans identified in the \textsc{ScotReg} corpus.}
    \label{tab:experiment}
\end{table}

Two domain experts annotated the 165 \textsc{Object} spans, see Table~\ref{tab:experiment}. The inter-annotator agreement was found to be Cohen \textit{k}=0,79. If we take the average of their judgement, this comes down to 116 (70,3\%) exact matches, 40 (23,6\%) partial matches, and 9 (5,5\%) non-objects. 
Examples of each labelled object and relevant parts of the sentence context include:
\begin{itemize}
    \item \textbf{Exact match}: `\textit{mechanical input air ventilation systems}' in ``\textit{Positive input systems - mechanical input air ventilation systems have been} [...]''.
    \item \textbf{Exact match}: `\textit{the warning}' in ``[...] \textit{the earliest possible warning} [...]'' -- where `\textit{earliest}' and `\textit{possible}' are modifiers that are not inherently part of the concept `\textit{warning}'.
    \item \textbf{Exact match}: `\textit{wall/roof junctions}' in ``[...] \textit{and at wall/roof junctions, wall/floor junctions and } [...]''.
    \item \textbf{Partial match}: `\textit{the control equipment}' in ``[...] \textit{the control and indicating equipment operates a fire alarm system} [...]'' -- `\textit{the control and indicating equipment}' should be treated as a single \textsc{Object} span.
    \item \textbf{Partial match}: `\textit{Articles 15 \& 16 }' in ``[...] \textit{implements the terms of Articles 15 \& 16 of Directive 2010/31/EU on} [...]'' -- should be split into `\textit{Articles 15}' and `\textit{Articles 16}'.
    \item \textbf{Partial match}: `\textit{primary}' in ``[...] \textit{educations centres, schools (nursery, primary, secondary, special)} [...]'' -- should identify the discontiguous part `\textit{school}'.
    \item \textbf{Not an object}: `\textit{sleeping}' in ``\textit{Rooms intended for sleeping should be} [...]'' -- should be an \textsc{Action} span here.
    \item \textbf{Not an object}: `\textit{land subject}' in ``[...] \textit{development may be given approval on land subject to} [...]'' -- the \textsc{Object} `\textit{land}' is modified by `\textit{subject to}'. 
    \item \textbf{Not an object}: `\textit{changes 1}' in ``\textit{Schedule 1 - changes to building types 1 and 20.}'' -- may be the result of overfitting on certain discontiguous patterns.
\end{itemize}

\subsection{Discussion}\label{e:discussion}
Despite the small size of \textsc{SPaR}.txt, the trained model discovers a large number (16K) of \textsc{Object} spans in 13K sentences. These spans cover most terms that are explicitly defined in the \textsc{ScotReg} corpus (89,84\%). The defined terms that the model did not identify are expressed in patterns that were never seen during training, although some do not occur verbatim in the texts. A significant proportion (70,3\%) of identified \textsc{Object} spans exactly match human judgement. Because annotation is cheap for our task, it is straightforward to create additional gold training samples and improve performance. To this end, partial matches can help identify phenomena that were not seen during training, e.g., `\textit{Articles 15 \& 16}' and `\textit{subject to}' as listed above. False positive \textsc{Object} spans provide insight in phenomena that the model currently overfits on, and may potentially help balance future iterations of our dataset. Moreover, the predicted outputs for the \textsc{ScotReg} corpus are valuable to the creation of a lexicon for \ACC.  

\section{Conclusions}\label{conclusions}
Regulatory documents are an important part of the legal framework, with research on \ACC methods focusing on the grand goal of semantic parsing. This study introduces a much simpler parsing task that requires few training examples, with the additional benefit that the collection of a dataset is cheap. We presented the small \textsc{SPaR}.txt dataset and trained a sequence tagger that can process single-word and multi-word spans. 
We showed that the \textsc{Object} spans identified in the \textsc{ScotReg} corpus cover most of the existing, limited set of defined terms. Moreover, the model achieves reasonable accuracy when it comes to discovering \textsc{Object} spans, regardless of whether these are discontiguous or not. 

The annotation of gold training data for the presented approach is cheap, because the annotation task is simple. But the results can benefit the research on \ACC, e.g., the output of our task may support more complicated semantic annotation tasks, \IE and \IR, as well as the development of a domain-specific lexicon. Future work will focus on clustering the identified spans to develop a semantic lexicon, balancing and growing the dataset, as well as using predicted outputs for \IR in support of manual \ACC. Finally, we will explore how well the presented approach performs in other domains with similar text characteristics.

\section*{Acknowledgements}
This research is funded by the IC3 (International Centre for Connected Construction) of Northumbria University. The authors are grateful to Julian Vincent for his thoughts on the annotation process. We would also like to congratulate Dr. Ben Trevett for passing his viva. 

\bibliography{references}
\bibliographystyle{acl_natbib}

{
    \newgeometry{}
    \appendix
    
\section{Overview of spans and tags for \textsc{SPaR}.txt and \textsc{ScotReg} predictions} \label{appendix}


\begin{table*}[ht]
\centering \small
    \begin{tabular}{rl|llllll|l}
                                  & \textbf{All} & \textbf{Train} & \%*    & \textbf{Development} & \%*    & \textbf{Test} & \%*                                            & \textbf{Predicted}  \\ \toprule
    Domestic                      & 98           & 62             & {\scriptsize63,27} & 19                   & {\scriptsize19,39} & 17            & {\scriptsize17,35}     & 6.313              \\
    Non-domestic                  & 102          & 58             & {\scriptsize56,86} & 21                   & {\scriptsize20,59} & 23            & {\scriptsize22,55}     & 7.293              \\
                                  &              &                & \textbf{}      &                      & \textbf{}      &               & \textbf{}                      & 13.606             \\ \arrayrulecolor{lightgray}\hline
    \textbf{avg. sent. (token$^{\dagger}$) length  }   & 27,96        & 28,31          & \textbf{}      & 28,05                & \textbf{}      & 26,83         & \textbf{} & 26,72              \\
    Std. dev                      & 15,31        & 15,50          & \textbf{}      & 14,03                & \textbf{}      & 15,92         & \textbf{}                      & 15,53              \\
    Shortest                      & 2            & 2              & \textbf{}      & 4                    & \textbf{}      & 2             & \textbf{}                      & 1                  \\
    Longest                       & 85           & 85             & \textbf{}      & 60                   & \textbf{}      & 63            & \textbf{}                      & 199                \\
                                  &              &                & \textbf{}      &                      & \textbf{}      &               & \textbf{}      &                    \\ \hline
    \textbf{Span types }          & 3.253        & 1.985          & {\scriptsize61,02} & 633                  & {\scriptsize19,46} & 635           & {\scriptsize19,52}     & 216.949          \\
    Discourse                     & 1.312        & 788            & {\scriptsize60,06} & 258                  & {\scriptsize19,66} & 266           & {\scriptsize20,27}     & 87.957            \\
    Object                        & 1.122        & 695            & {\scriptsize61,94} & 224                  & {\scriptsize19,96} & 203           & {\scriptsize18,09}     & 74.193            \\
    Action                        & 476          & 294            & {\scriptsize61,76} & 82                   & {\scriptsize17,23} & 100           & {\scriptsize21,01}     & 31.835           \\
    Functional                    & 343          & 206            & {\scriptsize60,64} & 69                   & {\scriptsize20,12} & 66            & {\scriptsize19,24}     & 22.964           \\
                                  &              &                & \textbf{}      &                      & \textbf{}      &               & \textbf{}      &                    \\
    Nr. discontiguous             & 193          & 122             & {\scriptsize63,21} & 33                   & {\scriptsize17,10} & 30            & {\scriptsize15,54}    &   6,450  {\scriptsize Objects}   \\
                                  &              &                & \textbf{}      &                      & \textbf{}      &               & \textbf{}                      & 2,007 {\scriptsize Actions}    \\ \hline
    \textbf{avg. span (char) length} & 7,51         & 7,49           & \textbf{}      & 7,49                 & \textbf{}      & 7,58          & \textbf{}                   & 7,14               \\
    Discourse                     & 2,56         & 2,50           & \textbf{}      & 2,53                 & \textbf{}      & 2,77          & \textbf{}                      & 2,44               \\
    Object                        & 12,54        & 12,41          & \textbf{}      & 12,40                & \textbf{}      & 13,11         & \textbf{}                      & 12,40              \\
    Action                        & 9,51         & 9,54           & \textbf{}      & 9,87                 & \textbf{}      & 9,15          & \textbf{}                      & 8,31               \\
    Functional                    & 7,22         & 7,08           & \textbf{}      & 7,29                 & \textbf{}      & 7,58          & \textbf{}                      & 6,53               \\
                                  &              &                & \textbf{}      &                      & \textbf{}      &               & \textbf{}      &                    \\ \hline\arrayrulecolor{black}
    \textbf{Tag types }                    &              &                & \textbf{}      &                      & \textbf{}      &               & \textbf{}      &                    \\ 
    IH-obj    & 1.551 & 931     & {\scriptsize60,03}     & 336   &  {\scriptsize21,66}     & 284   &  {\scriptsize18,31}                                                & 108.507 \\
    BH-dis    & 1.312 & 788     & {\scriptsize60,06}     & 258   &  {\scriptsize19,66}     & 266   &  {\scriptsize20,27}                                                & 87.957  \\
    BH-obj    & 1.071 & 659     &  {\scriptsize61,53}     & 219   &  {\scriptsize20,45}     & 193   &  {\scriptsize18,02}                                                & 74.311  \\
    BH-act    & 472   & 292     &  {\scriptsize61,86}     & 81    &  {\scriptsize17,16}     & 99    &  {\scriptsize20,97}                                                & 31.843  \\
    BH-func   & 343   & 208     &  {\scriptsize60,64}     & 69    & {\scriptsize 20,12}     & 66    &  {\scriptsize19,24}                                                & 22.964  \\
    IH-act    & 331   & 205     &  {\scriptsize61,93}     & 58    &  {\scriptsize17,52}     & 68    &  {\scriptsize20,54}                                                & 19.013\\
    BD-obj    & 149   & 94      &  {\scriptsize63,09}     & 25    &  {\scriptsize16,78}     & 30    &  {\scriptsize20,13}                                                & 6.450   \\
    IH-func   & 136   & 88      &  {\scriptsize64,71}     & 30    &  {\scriptsize22,06}     & 18    &  {\scriptsize13,24}                                                & 6.784   \\
    ID-obj    & 134   & 88      &  {\scriptsize65,67}     & 31    &  {\scriptsize23,13}     & 15    &  {\scriptsize11,19}                                                & 3.114  \\
    IH-dis    & 45    & 16      &  {\scriptsize35,56}     & 7     &  {\scriptsize15,56}     & 22    &  {\scriptsize48,89}                                                & 482     \\
    BD-act    & 34    & 20      &  {\scriptsize58,82}     & 6     &  {\scriptsize17,65}     & 8     &  {\scriptsize23,53}                                                & 2.007   \\
    ID-act    & 14    & 8       &  {\scriptsize57,14}     & 2     &  {\scriptsize14,29}     & 4     &  {\scriptsize28,57}                                                & 74   \\ \bottomrule
\end{tabular}
 \caption{Overview of sentence and span statistics for  \textsc{SPaR}.txt (Gold/Train/Dev/Test), as well as for the predictions (Predictions) over the entire \textsc{ScotReg} corpus. \%* indicates the percentage of all gold data. $\dagger$ Note that BERT tokenization is used here.}
    \label{tab:overview_spans}
\end{table*}

\clearpage
\section{Overview of examples} \label{examples_overviewe}
{\centering \footnotesize
\begin{longtable}{p{0.04\linewidth}p{0.60\linewidth}p{0.18\linewidth}p{0.08\linewidth}} \\
 
Sect.           & Full sentence          & \multicolumn{2}{l}{Source}      \\ \toprule \endhead
\multirow{2}{*}{\ref{mwe:problem_statement} }          & \textbf{A roof covering or roof light which forms part of an internal ceiling lining should} also follow the guidance to Standard 2.5 Internal linings.                                                                                                                                                                                            & {\footnotesize{2.8.0 Introduction}}           & Domestic     \\ \hline
\ref{mwe:task}       & \textbf{The Building (Scotland) Act 2003 gives Scottish Ministers} the power to make building regulations to:                                                                                                                                                                                                                                                                     & {\footnotesize{0.1.1 Introduction}}                                        & Domestic     \\ \arrayrulecolor{lightgray}\cline{2-4}\arrayrulecolor{black}\arrayrulecolor{black}
                                                          & In order to allow unobstructed access to a domestic building for fire and rescue service personnel\textbf{, a paved (or equivalent) footpath at least 900mm wide }(see also Section 4 Safety) should be provided to the normal entrances, of a building.                                                                                                                          & {\footnotesize{2.12.4 Access for fire and rescue service personnel}}       & Domestic     \\ \hline
\ref{mwe:assumptions} & Section 6 Energy, indicates that less demanding U-values can be adopted for the \textbf{insulation envelope} of certain types of limited life buildings, other than dwellings and residential buildings.                                                                                                                                                                          & {\footnotesize{0.6.1 Explanation}}                                        & Domestic     \\  \arrayrulecolor{lightgray}\cline{2-4}\arrayrulecolor{black}\arrayrulecolor{black}
                                                          & In conversions for example, it may be easier to install \textbf{self-contained emergency luminaries} than to install a protected circuit to the existing lighting system                                                                                                                                                                                                          & {\footnotesize{2.10.2 Protected circuits}}                                 & Domestic     \\ \hline
\ref{mwe:data}       & To enable the continued use of existing stocks of  building modules and sub-assemblies, subject to fabric insulation meeting the U-values noted in clause 6.C.3, a modifying factor can be applied to increase \textbf{the Target Emissions Rating} (TER) for the building.                                                                                                       & {\footnotesize{n\_6.C257\_i0\_s\_0}}                                      & Non-domestic \\  \arrayrulecolor{lightgray}\cline{2-4}\arrayrulecolor{black}\arrayrulecolor{black}
                                                          & This is termed `\textbf{offensive fire-fighting}' and is normal practice regardless of whether people are in the building or not.                                                                                                                                                                                                                                                 & {\footnotesize{d\_2.14.0\_i4\_s\_1}}                                      & Domestic     \\ \arrayrulecolor{lightgray}\cline{2-4}\arrayrulecolor{black}\arrayrulecolor{black}
                                                          & \textbf{BS 8000-15: 1990} - Workmanship on building sites                                                                                                                                                                                                                                                                                                                         & {\footnotesize{d\_0.8.8\_i3\_\#1\_s\_38}}                                  & Domestic     \\ \arrayrulecolor{lightgray}\cline{2-4}\arrayrulecolor{black}\arrayrulecolor{black}
                                                          & The minimum 3m separation in the diagram below \textbf{should be maintained} between each 5m$^{2}$ panel.                                                                                                                                                                                                                                                                              & {\footnotesize{d\_2.5.7\_i5\_\#1\_s\_0}}                                   & Domestic     \\ \arrayrulecolor{lightgray}\cline{2-4}\arrayrulecolor{black}\arrayrulecolor{black}
                                                          & The guidance in this clause \textbf{takes account of} the audibility levels in adjoining rooms and \textbf{the effect of smoke travelling along a ceiling.}  & {\footnotesize{d\_2.11.7\_i0\_s\_0}}                                   & Domestic     \\ \arrayrulecolor{lightgray}\cline{2-4}\arrayrulecolor{black}\arrayrulecolor{black}
                                                          & The guidance in this clause \textbf{takes account of} the audibility levels in adjoining rooms and the effect of smoke travelling along a ceiling.                                                                                                                                                           & {\footnotesize{d\_2.11.7\_i0\_s\_0}}                                       & Domestic     \\ \arrayrulecolor{lightgray}\cline{2-4}\arrayrulecolor{black}\arrayrulecolor{black}
                                                          & CO2 monitoring equipment should be provided in the apartment expected to be \textbf{the main or principal bedroom} in a dwelling where infiltrating air rates are less than 15m$^{3}$/hr/m$^{2}$ @ 50 Pa.                                                                                                                                                       & {\footnotesize{d\_3.14.2\_i1\_s\_0}}                                       & Domestic     \\ \arrayrulecolor{lightgray}\cline{2-4}\arrayrulecolor{black}\arrayrulecolor{black}
                                                          & Non-domestic use within dwellings - accommodation \textbf{up to} 50m$^{2}$ used by an occupant of a dwelling in their professional or business capacity should be considered as a part of the dwelling.                                                                                                                                                                  & {\footnotesize{d\_6.9.1\_i3\_s\_0}}                                        & Domestic     \\ \arrayrulecolor{lightgray}\cline{2-4}\arrayrulecolor{black}\arrayrulecolor{black}
                                                          & In the measurement of height or depth from ground which \textbf{is not level} the height or depth shall be taken to be the mean height or depth, except that:    for the purpose of types 1, 2, 3, 4, 5, 18 or 19 of schedule 3, and for any other purpose where the difference in level is more than 2.5m the height or depth shall be taken to be the greatest height or depth. & d\_0.7.2\_i1\_\#4\_s\_0                                   & Domestic     \\ \arrayrulecolor{lightgray}\cline{2-4}\arrayrulecolor{black}\arrayrulecolor{black}
                                                          & Other pipes should be capped at both ends and at any point of connection, to ensure rats \textbf{cannot gain} entry.                                                                                                                                                                                                                                                              & {\footnotesize{n\_3.5.5\_i1\_s\_1}}                                        & Non-domestic  \\ \arrayrulecolor{lightgray}\cline{2-4}\arrayrulecolor{black}\arrayrulecolor{black}
                                                          & In order to allow unobstructed access to a domestic building for \textbf{fire and rescue service personnel}, a paved (or equivalent) footpath at least \textbf{900mm wide} (see also Section 4 Safety) should be provided to the normal entrances, of a building.                                                                                                                          & {\footnotesize{d\_2.12.4\_i1\_s\_0}}                                       & Domestic   \\ \arrayrulecolor{lightgray}\cline{2-4}\arrayrulecolor{black}\arrayrulecolor{black}
                                                          & The layout of a dwelling,\textbf{ the size and orientation of the windows}, the thermal mass, level of insulation, airtightness, and ventilation can have a significant affect on the demand for heat.                                                                                                                                                                            & {\footnotesize{d\_3.13.1\_i0\_s\_0}}                                       & Domestic     \\ \arrayrulecolor{lightgray}\cline{2-4}\arrayrulecolor{black}\arrayrulecolor{black}
                                                          & Standard 7.1 - amendments have been made to guidance with regard to the carbon dioxide (CO2) emissions target within \textbf{the Silver and Gold level} of Sustainability labelling in relation to the CO2 emissions target introduced by the 2015 energy standards.                                                                                                              & {\footnotesize{d\_7.0.5\_i1\_\#0\_s\_0}}                                   & Domestic     \\ \arrayrulecolor{lightgray}\cline{2-4}\arrayrulecolor{black}\arrayrulecolor{black}
                                                          & Materials that are susceptible to changes in their properties may be used in building work and will meet the requirements of the regulations if the residual properties, including\textbf{ the structural properties}:                                                                                                                                                            & {\footnotesize{d\_0.8.7\_i1\_s\_0}}                                        & Domestic \\ \arrayrulecolor{lightgray}\cline{2-4}\arrayrulecolor{black}\arrayrulecolor{black}
                                                          & The collation and dissemination of information relating to \textbf{matters of structural concern} is a vital element of achieving safe structures.                                                                                                                                                            & {\footnotesize{d\_1.0.1\_i2\_s\_2}}                                        & Domestic  \\  \hline
\ref{e:objects}     & Positive input systems - \textbf{mechanical input air ventilation systems} have been successfully installed in existing dwellingswith the objective of overcoming problems of surface condensation and mould growth.                                                                                                                                                              & {\footnotesize{3.14.11 Mechanical ventilation and systems}}                & Domestic     \\ \arrayrulecolor{lightgray}\cline{2-4}\arrayrulecolor{black}\arrayrulecolor{black}
                                                          & This is to give occupants and staff \textbf{the earliest possible warning} of an outbreak of fire and allow time for assisting occupants in an emergency to evacuate the building or for horizontal progressive evacuation initially to an adjacent sub-compartment which leads to a compartment exit.                                                                            & {\footnotesize{2.11.5 Hospitals}}                                          & Non-domestic \\ \arrayrulecolor{lightgray}\cline{2-4}\arrayrulecolor{black}\arrayrulecolor{black}
                                                          & These `bridges' commonly occur around openings such as lintels, jambs and sills and at \textbf{wall/roof junctions}, wall/floor junctions and where internal walls penetrate the outer fabric.                                                                                                                                                                                    & {\footnotesize{3.15.4 Surface condensation - thermal bridging}}            & Non-domestic \\ \arrayrulecolor{lightgray}\cline{2-4}\arrayrulecolor{black}\arrayrulecolor{black}
                                                          & Normally \textbf{the control and indicating equipment }operates a fire alarm system and it may perform other signalling or control functions as well.                                                                                                                                                                                                                             & {\footnotesize{2.11.3 Categories of fire detection and fire alarm system}} & Non-domestic \\ \arrayrulecolor{lightgray}\cline{2-4}\arrayrulecolor{black}\arrayrulecolor{black}
                                                          & This regulation implements the terms of \textbf{Articles 15 \& 16} of Directive 2010/31/EU on the Energy Performance of Buildings (EPBD).                                                                                                                                                                                                                                         & {\footnotesize{0.17.1 Explanation}}                                        & Domestic     \\ \arrayrulecolor{lightgray}\cline{2-4}\arrayrulecolor{black}\arrayrulecolor{black}
                                                          & • education centres, \textbf{schools (nursery, primary, secondary, special) }                                                                                                                                                                                                                                                                                                     & {\footnotesize{6.9.3 Location of an energy performance certificate}}      & Non-domestic \\ \arrayrulecolor{lightgray}\cline{2-4}\arrayrulecolor{black}\arrayrulecolor{black}
                                                          & Rooms intended for \textbf{sleeping} should be separated by a door that will act as a sound barrier and reduce noise transference.                                                                                                                                                                                                                                                & {\footnotesize{5.2.5 Doors in internal walls}}                             & Non-domestic \\ \arrayrulecolor{lightgray}\cline{2-4}\arrayrulecolor{black}\arrayrulecolor{black}
                                                          & Pressure for land development may mean that development may be given planning approval on \textbf{land subject} to some risk of flooding.                                                                                                                                                                                                                                         & {\footnotesize{3.3.0 Introduction}}                                        & Domestic     \\ \arrayrulecolor{lightgray}\cline{2-4}\arrayrulecolor{black}\arrayrulecolor{black}
                                                          & • Schedule 1 - \textbf{changes to building types 1} and 20.                                                                                                                                                                                                                                                                                                                       & {\footnotesize{0.2.1 Explanation of Regulation 1}}                        & {\footnotesize{Non-domestic}} \\ \bottomrule
    \caption{Complete overview of the examples used in the paper  and their location in \textsc{ScotReg}. Examples that occur in \textsc{SPaR}.txt have an identifier, such as n\_6.C257\_i0\_s\_0. These identifiers are created as follows: domestic (d) or non-domestic (n), the section (6.C2) -- in this case concatenated with an integer (57) that is incremented each time a section name is reused, the item index (0 for first item) that indicates which item in the list of this section, and the sentence number (s\_0 for first sentence).`\textit{\#}' in the identifier means that the sentence occurs in a list or table, with lists and tables 0-indexed for each section. 
    }
    \label{tab:examples_overview}
\end{longtable}
}
    \restoregeometry
}

\end{document}